\documentclass[letterpaper, 10 pt, conference]{ieeeconf}  
\IEEEoverridecommandlockouts                             
\usepackage{graphics} 
\usepackage{epsfig} 
\usepackage{times} 
\usepackage{amsmath} 
\usepackage{amssymb}  
\usepackage{xcolor}
\usepackage{bm}
\usepackage{cancel}
\usepackage{cuted}
\usepackage{graphicx} 
\usepackage{color}
\usepackage{setspace}
\usepackage{wrapfig}
\usepackage{tikz}
\usetikzlibrary{shapes,arrows.meta,calc}
\usetikzlibrary{positioning}
\usepackage{caption}
\usepackage{subcaption}
\usepackage{booktabs}
\usepackage{multirow}
\usepackage{multicol}
\usepackage{soul}
\usepackage{cite}
\usepackage[normalem]{ulem}
\usepackage{xcolor}
\usepackage{mathtools}
\usepackage{stfloats}

\let\labelindent\relax 
\usepackage{enumitem}
\usepackage{tabularx}
\usepackage{courier}
\usepackage{algorithm}
\usepackage{algorithmic}
\usepackage{makecell}
\usepackage{url}
\usepackage{units}
\usepackage{placeins}
\usepackage{arydshln} 
\usepackage{hyperref}

\usepackage{pifont}
\usepackage{url}

\newcommand{\todocite}[1]{\textcolor{red}{[TODO(cite)]}}

\allowdisplaybreaks

\title{\LARGE 
\textbf{
GRaD-Nav: Efficiently Learning Visual Drone \underline{Nav}igation with \underline{G}aussian \underline{Ra}diance Fields and \underline{D}ifferentiable Dynamics
}
}

\author{Qianzhong Chen$^{1}$, Jiankai Sun$^{2}$, Naixiang Gao$^{1}$, JunEn Low$^{1}$, Timothy Chen$^{2}$, and Mac Schwager$^{2}$
\thanks{$^{1}$ Authors are with the Department of Mechanical Engineering, Stanford University, Stanford, CA 94305, USA. 
    {\tt\footnotesize \{qchen23,ngao4,jelow\}@stanford.edu}}
\thanks{$^{2}$ Authors are with the Aeronautics and Astronautics Department, Stanford University, Stanford, CA 94305, USA.
    {\tt\footnotesize \{jksun,schwager,chengine\}@stanford.edu}}%
    \thanks{This work was supported in part by ONR grant N00014-23-1-2354 and a gift from Meta.
    Toyota Research Institute provided funds to support this work.}
    \thanks{\textit{(Corresponding author: Q. Chen)}}
}

\begin{document}
\maketitle
\thispagestyle{empty}
\pagestyle{empty}

\begin{abstract}
Autonomous visual navigation is an essential element in robot autonomy. Reinforcement learning (RL) offers a promising policy training paradigm. However, existing RL methods suffer from high sample complexity, poor sim-to-real transfer, and limited runtime adaptability. These problems are particularly challenging for drones, with complex nonlinear and unstable dynamics, and strong dynamic coupling between control and perception. In this paper, we propose a novel framework that integrates 3D Gaussian Splatting (3DGS) with differentiable deep reinforcement learning (DDRL) to train vision-based drone navigation policies. By leveraging high-fidelity 3D scene representations and differentiable simulation, our method improves sample efficiency and sim-to-real transfer. Additionally, we incorporate a \underline{C}ontext-aided \underline{E}stimator \underline{Net}work (CENet) to adapt to environmental variations at runtime. Moreover, by curriculum training in a mixture of different surrounding environments, we achieve in-task generalization, the ability to solve new instances of a task not seen during training. Drone hardware experiments demonstrate our method's high training efficiency compared to state-of-the-art RL methods, zero shot sim-to-real transfer for real robot deployment without fine tuning, and ability to adapt to new instances within the same task class (e.g. to fly through a gate at different locations with different distractors in the environment). Our simulator and training framework are open-sourced at: \url{https://github.com/Qianzhong-Chen/grad_nav}.
    
\end{abstract}


\section{Introduction}

Autonomous drones have the potential to improve practices in agriculture, environmental management, and search and rescue \cite{ahirwar2019application, daud2022applications, fan2019applications}. A critical aspect of these robots' functionality is their efficient navigation and control in complex, unstructured environments \cite{zhou2022swarm}. Traditional approaches to this problem have predominantly relied on a stack of different modules including perception, localization, mapping, planning, and control \cite{hanover2024autonomous, floreano2015science, chen2023simultaneous}. However, the integration of these different modules has many issues, including high system complexity and computational overhead, communication latency between modules, multiple points of failure, and difficult-to-characterize error propagation between modules. 

Recent advancements in machine learning have offered alternative solutions to these challenges. Deep reinforcement learning (DRL) has been proven to be successful in drone racing \cite{kaufmann2023champion}. By leveraging DRL, robots can potentially learn to directly map sensor inputs to control outputs, bypassing the need for explicit modular separation \cite{wu2024whole}. While promising, so far, these DRL solutions have been mostly limited to drone racing-like tasks but have not proved successful in navigating complex and unstructured environments. To tackle the challenge, one of the most important bottlenecks lies on the difficulty in getting high-quality perception data when training the policy in conventional simulators \cite{todorov2012mujoco, coumans2016pybullet}. Synthetic visual data still struggles to accurately capture fine details, often leading to a substantial sim-to-real gap~\cite{pan2020cross}.

\hfill
\begin{figure}[t]
    \flushright
    \includegraphics[width=\linewidth]{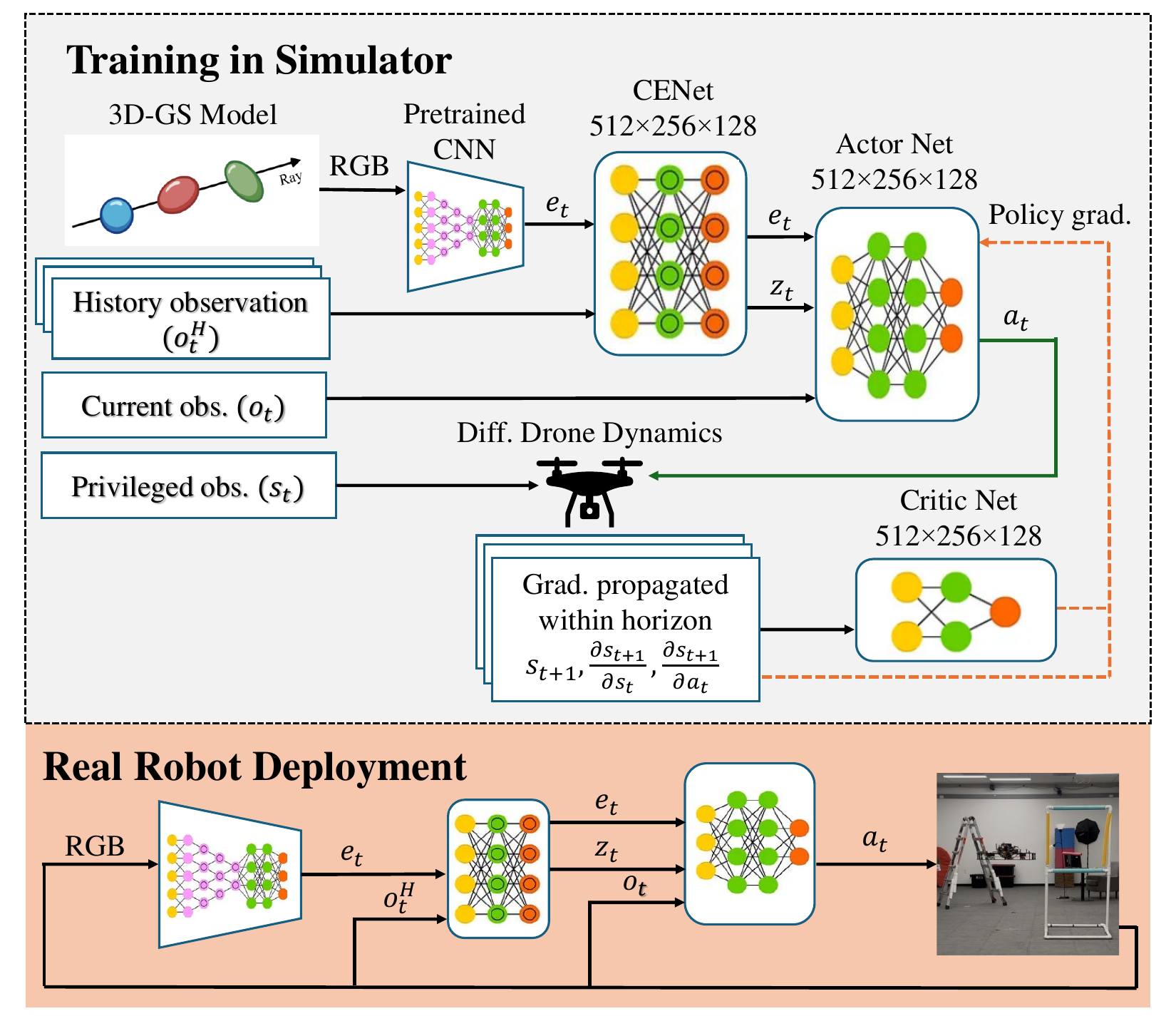} 
    \captionof{figure}{Our GRaD-Nav architecture combines a visual+dynamics context encoder (CENet) within an Actor-Critic framework, trained end-to-end using a differentiable drone dynamics model and 3D Gaussian Splatting scene representation for photo-realistic visuals at training time. The policy transfers zero-shot to drone hardware and adapts to new navigation task instances at runtime.}
    \label{fig:model_struc}
\end{figure}

3D reconstruction has demonstrated exceptional capability in generating high-fidelity 3D scene representations from sparse views~\cite{mildenhall2021nerf, kerbl2023gaussiansplatting}, making it an attractive approach for creating an environment model~\cite{shorinwa2024fast,shorinwa2025siren,sun2023nerf}. Flying in 3D reconstruction has been proven successful for multiple drone tasks including trajectory planning, pose estimation, and visual navigation \cite{adamkiewicz2022vision, chen2024splat, chen2024catnips, yen2021inerf, tagliabue2023tube, quach2024gaussian}. Moreover, the recent work SOUS VIDE \cite{low2024sous} utilizes Flying in a Gaussian Splat (FiGS) to train end-to-end drone navigation policies, and demonstrated successful sim-to-real transfer. However, the works above are based on imitation learning, which requires a large amount of high-quality expert pilot data, long training time, and suffers from a lack of generalization to new task instances.

On the other hand, differentiable simulators \cite{hu2019difftaichi, macklin2022warp, freeman2021brax, howell2022dojo} and differentiable deep reinforcement learning (DDRL) algorithms \cite{mora2021pods}, \cite{xu2022accelerated} alleviate another critical issue of sample efficiency for conventional DRL. Differentiable simulators not only provide the simulated object's next state, but also delivers the gradient of the simulated state with respect to the action and previous state, enhancing model training with gradient knowledge rather than simply trial and error. By leveraging a custom smoothing critic function and truncated learning window, the Short Horizon Actor-Critic (SHAC) algorithm \cite{xu2022accelerated} alleviates the problems of local minima and exploding/vanishing gradients that commonly occur in DDRL training.

Even with exteroceptive sensors like RGB cameras, it is still a significant challenge for drone policies to understand the complex surrounding environment, especially the spatial relationship between drone and obstacle. 
To address this, we introduce a context-aided estimator network (CENet)~\cite{nahrendra2023dreamwaq} positioned before the policy network, enabling the agent to infer crucial environmental properties at runtime. While previous approaches have primarily relied on proprioceptive sensing, we extend this idea by including RGB camera sensor inputs, leading to improved in-task adaptability in navigation scenarios. Concretely, CENet outputs a latent embedding at runtime summarizing visual and dynamical context. The policy is conditioned on this code, enabling the drone to modify its behavior in response to the sensed runtime environment.

To achieve the goal of visual-motor navigation, we propose a novel approach that leverages 3DGS in conjunction with DDRL, using SHAC-like training algorithm and a policy architecture augmented with a CENet module for in-context runtime adaptation. By utilizing this combination, we develop a visual navigation policy learning framework that transfers to real-world drone navigation performance. Our main contributions are:
\begin{itemize}
\item We introduce a simulator for training robot vision-based control policies by integrating 3DGS for high-fidelity visuals with a differentiable dynamics model to enable end-to-end gradient computation.
\item We propose GRaD-Nav, a DDRL algorithm designed for efficient end-to-end visual navigation policy learning, achieving higher sample efficiency than previous approaches.
\item We demonstrate in-task generalization by training a single policy for the drone to fly through gates at different locations with different distractors in the environment not seen at training time.
\end{itemize}

\section{Background} 
\label{sec:background}

\subsection{Differentiable Simulation and Differentiable Reinforcement Learning}
\subsubsection{Differentiable Dynamics Simulation}
Conceptually, we view the simulator as a differentiable abstract function, $\mathbf{s}_{t+1}=\mathcal{F}\left(\mathbf{s}_t, \mathbf{a}_t\right)$, which transitions a state $\mathbf{s}$ from time $t$ to $t+1$. Here, $\mathbf{a}_t$ represents the control input during that time-step. Given a differentiable scalar loss function $\mathcal{L}$ and its adjoint $\mathcal{L}^*=\frac{\partial \mathcal{L}}{\partial \mathbf{s}_{t+1}}$, the backward pass of the simulator computes:
\begin{equation}
\small
    \begin{aligned}
    \frac{\partial \mathcal{L}}{\partial \mathbf{s}_t}=\left(\frac{\partial \mathcal{L}}{\partial \mathbf{s}_{t+1}}\right)\left(\frac{\partial \mathcal{F}}{\partial \mathbf{s}_t}\right), \quad \frac{\partial \mathcal{L}}{\partial \mathbf{a}_t}=\left(\frac{\partial \mathcal{L}}{\partial \mathbf{s}_{t+1}}\right)\left(\frac{\partial \mathcal{F}}{\partial \mathbf{a}_t}\right).
    \end{aligned}
\end{equation}

\subsubsection{Short-Horizon Actor-Critic}
The Short-Horizon Actor-Critic method (SHAC) \cite{xu2022accelerated} was introduced to address the challenges associated with gradient-based policy learning. This method simultaneously trains a policy network (actor) $\pi_\theta$ and a value network (critic) $V_\phi$ during task execution, dividing the entire task horizon into several smaller sub-windows across learning episodes. 
Differentiable simulation allows for backpropagation of the gradient through states and actions within the sub-windows, providing an accurate policy gradient. 



The SHAC method operates in an on-policy manner as follows: In each learning episode, the algorithm samples $N$ trajectories $\left\{\tau_i\right\}$ of short horizon $h \ll H$ in parallel from the simulation, continuing from the end states of the trajectories in the previous learning episode. The policy loss is then computed as follows:
\begin{equation}
\scriptsize
    \begin{aligned}
\mathcal{L}_\theta=-\frac{1}{N h} \sum_{i=1}^N\left[\left(\sum_{t=t_0}^{t_0+h-1} \gamma^{t-t_0} \mathcal{R}\left(\mathbf{s}_t^i, \mathbf{a}_t^i\right)\right)+\gamma^h V_\phi\left(\mathbf{s}_{t_0+h}^i\right)\right],
    \end{aligned}
\end{equation}
In this context, $\mathbf{s}_t^i$ and $\mathbf{a}_t^i$ represent the state and action at step $t$ of the $i$-th trajectory, and $\gamma < 1$ is a discount factor used to stabilize the training process. Special handling, such as resetting the discount ratio, is applied when task termination occurs during trajectory sampling.

To compute the gradient of the policy loss $\frac{\partial \mathcal{L}_\theta}{\partial \theta}$, the simulator is integrated as a differentiable layer within the PyTorch \cite{paszke2019pytorch} computation graph, allowing for regular backpropagation. 
The differentiable simulator is crucial here as it enables full utilization of the underlying dynamics connecting states and actions, optimizing the policy to achieve better short-horizon rewards as well as trajectory's overall performance. After updating the policy $\pi_\theta$, the trajectories collected in the current learning episode are used to train the value function $V_\phi$. The value function network is trained using the following mean squared error (MSE) loss:
\begin{equation}
    \begin{aligned}
\mathcal{L}_\phi=\underset{\mathbf{s} \in\left\{\tau_i\right\}}{\mathbf{E}}\left[\left\|V_\phi(\mathbf{s})-\tilde{V}(\mathbf{s})\right\|^2\right],
    \end{aligned}
\end{equation}
where $\tilde{V}(\mathbf{s})$ is the estimated value of state $\mathrm{s}$, and is computed from the sampled short-horizon trajectories through a td-$\lambda$ formulation \cite{sutton1998reinforcement}.

\subsection{3D Gaussian Splatting}
3D Gaussian Splatting (3DGS) is a technique for representing and rendering 3D scenes by leveraging a continuous and compact set of anisotropic 3D Gaussian primitives. Introduced by Kerbl et al.~\cite{kerbl2023gaussiansplatting}, this method models scenes with a collection of Gaussians, each parameterized by a position $\mathbf{\mu}_i \in \mathbb{R}^3$, an anisotropic covariance matrix $\mathbf{\Sigma}_i \in \mathbb{R}^{3 \times 3}$, a color $\mathbf{c}_i \in \mathbb{R}^3$, and a opacity $\alpha_i \in [0,1]$. These parameters enable 3DGS to represent geometry and radiance efficiently while achieving real-time rendering.

In 3DGS, rendering involves projecting the Gaussians onto a 2D image plane and aggregating their contributions to approximate pixel colors. For a pixel $\mathbf{p}$, the accumulated radiance is computed as:
\begin{equation}
\mathbf{C}(\mathbf{p}) = \sum_{i=1}^N w_i(\mathbf{p}) \cdot \mathbf{c}_i,
\end{equation}
where $w_i(\mathbf{p})$ is the weight of the $i$-th Gaussian's contribution to $\mathbf{p}$. This weight is determined by the Gaussian distribution:
\begin{equation}
w_i(\mathbf{p}) = \mathcal{N}(\mathbf{p}; \mathbf{\mu}_i, \mathbf{\Sigma}_i) \cdot T_i,
\end{equation}
where $\mathcal{N}(\mathbf{p}; \mathbf{\mu}_i, \mathbf{\Sigma}_i)$ represents the 2D projected Gaussian, and $T_i$ is the transmittance term that handles occlusions. The transmittance is recursively defined as:
\begin{equation}
T_i = \prod_{j=1}^{i-1} (1 - \alpha_j),
\end{equation}
where $\alpha_j$ is the opacity of the $j$-th Gaussian.

Overall, 3DGS achieves real-time rendering with high visual fidelity, largely thanks to its point-based rasterization pipeline and the compactness of the Gaussian representation.





\section{Method}
We present a hybrid vision-differentiable simulation framework that combines 3DGS with a differentiable simulator to train robot visuomotor policies efficiently. At its core, we introduce GRaD-Nav, a DDRL algorithm tailored for end-to-end visual navigation, improving sample efficiency over prior methods. By training a unified policy across diverse 3DGS environments, our approach enhances task-level generalization, paving the way for more adaptable and autonomous mobile robots.

\subsection{Simulator Setting}

\subsubsection{Differentiable Quadrotor Dynamics Simulation}
\label{section:method_simulator}
We implemented a parallelized, differentiable quadrotor dynamics simulator in PyTorch that computes gradients through full state transitions. Our system takes body rates $\bm{\omega}^d_t \in \mathbb{R}^3$ and normalized thrust $c_t \in [0,1]$ as control inputs, and outputs the next state $\bm{s}_{t+1} = (\bm{p},\bm{q},\bm{v}, \bm{\omega}, \bm{a}) \in \mathbb{R}^{16}$ containing position, orientation (quaternion), linear velocity, angular velocity, and acceleration. The dynamics model is fully differentiable, with $\partial\bm{s}_{t+1}/\partial\bm{s}_t$ and $\partial\bm{s}_{t+1}/\partial(\bm{\omega}^d_t,c_t)$ computed automatically via PyTorch's computation graph.

The core dynamics are governed by rigid-body mechanics and a PD attitude controller. For a quadrotor with mass $m$ and inertia matrix $\bm{I}$, the angular acceleration $\dot{\bm{\omega}}$ is computed as:
\begin{equation}
    \bm{\tau} = \bm{K}_p(\bm{\omega}^d - \bm{\omega}) - \bm{K}_d\,\bm{\dot{\omega}},
\end{equation}
\begin{equation}
    \dot{\bm{\omega}} = \bm{I}^{-1}[\bm{\tau} - \bm{\omega} \times (\bm{I}\bm{\omega})].
\end{equation}
Orientation is represented by a unit quaternion $\bm{q} \in \mathbb{S}^3 \subset \mathbb{R}^4$, which is updated through quaternion integration:
\begin{equation}
    \bm{q}_{t+1} = \text{norm}\left(\bm{q}_t + \frac{\Delta t}{2}\bm{q}_t \otimes [0\ \bm{\omega}]^\top\right),
\end{equation}
where $\otimes$ denotes quaternion multiplication. Linear acceleration in world frame is derived from rotation matrix $\bm{R}(\bm{q})$ and thrust $T = c\cdot T_{\max}$:
\begin{equation}
    \bm{a} = \frac{1}{m}\bm{R}(\bm{q})[0\ 0\ T]^\top + \bm{g}.
\end{equation}
The state $\bm{s}_t = [\bm{p}_t, \bm{v}_t, \bm{q}_t, \bm{\omega}_t]$ consists of position, velocity, orientation (quaternion), and angular velocity. The control input $\bm{u}_t = (\bm{\omega}^d_t, c_t)$ includes desired body rates and normalized thrust. To achieve more accurate dynamic simulation, we also incorporated motor delay and drag force modeling in the simulator.

\subsubsection{Hybrid simulation with 3DGS}
We used a pre-trained 3DGS model to deliver the drone's first person perspective visual information and to imitate the drone's onboard RGB camera's input during real flight in the same environment. Every step, the differentiable dynamic simulator sends the batched drone poses $\mathbf{T} = (\bm{p}, \bm{q}) \in \mathbb{R}^7$ to the 3DGS model and renders simulated RGB images. 

Meanwhile, 3DGS also provides us with a ready-made point cloud model for the same environment, which can be used to set up reward waypoints and plan a reference trajectory for guiding the DDRL training, as well as conducting collision checks during simulation. More details will be introduced in Section \ref{sec:nav_reward}.

\subsubsection{Implementation datails}
We used a standard open-source codebase, Nerfstudio \cite{tancik2023nerfstudio} as our 3DGS training and inference platform. The differentiable drone dynamics model is also implemented with PyTorch, which enables efficient Jacobian computation through autograd for training the policy using our GRaD-Nav algorithm. When parallel simulating 128 drones flying in a highly unstructured and cluttered area (room size $\approx 100 \, m^2$, 3DGS model size $\approx$ 1.5GB), setting the simulated time step as 0.05s (20Hz), the simulation wall-clock time is about 0.07s per step, tested on a desktop with i9-13900K CPU, RTX 4090 GPU, and 48GB RAM. The detailed account of each item's time consumption during simulation is listed in Table \ref{table:simulation_time}.

\begin{table}[h!]
\small
\centering
\caption{Computation time consumption ratio of different items of hybrid differentiable-3DGS simulator for each step}
\begin{tabular}{cc}
\toprule
\textbf{Item}                      & \textbf{Ratio (\%)} \\
\hline 
Diff. Dynamics Sim. & 33.5                                 \\
3DGS Rendering                    & 55.7                                 \\
Collision Check        & 10.8 \\
\bottomrule
\end{tabular}
\label{table:simulation_time}
\end{table}

\subsection{Model Architecture}
\subsubsection{Perception network}
We utilize a pretrained SqueezeNet \cite{iandola2016squeezenet} convolutional neural network (CNN) to process the RGB perception information of the drone from both 3DGS and onboard camera during training time and deployment time, respectively. SqueezeNet is a light-weighted, efficient, and robust CNN that can extract important information from drone's perception in real-time. Based on this, we finetuned a fully connected layer to downsample SqueezeNet's raw output from 512 to 24 dimensions to get the visual information embedding $\mathbf{e}_t \in \mathbb{R}^{24}$. This embedding is used to train policy network and context-aided estimator network along with the observation.

\subsubsection{Policy network}
The policy network $\pi_{\theta}(\mathbf{a}_{t+1}|\mathbf{o}_t, \mathbf{z}_t, \mathbf{e}_t)$ is a 3-layer multilayer perceptron (MLP) parameterized by $\theta$, each layer has 512, 256, 128 neurons, where  $\mathbf{z}_t \in \mathbb{R}^{16}$ is the latent embedding encoded by CENet and $\mathbf{o}_t \in \mathbb{R}^{16}$ is the observation defined as following:
\begin{equation}
    \mathbf{o}_t = \begin{bmatrix} \bm{h}_t & \bm{q}_t & \bm{v}_t & \bm{a}_t & \bm{a}_{t-1}  \end{bmatrix}^{T},
\end{equation}
where $\bm{h}_t$, $\bm{q}_t$, $\bm{v}_t$ are drone body's height, quaternion, and linear velocity, respectively. $\bm{a}_t$ and $\bm{a}_{t-1}$ are current action and previous action. We do not need to explicitly estimate drone's x-y positions. \emph{All variables in the observation can be obtained from the drone's onboard sensors, which makes it possible to achieve 0-shot sim-to-real transfer.} 


\subsubsection{Value network}
The value network $V_{\phi}(\mathbf{s}_t, \mathbf{z}_t)$ is parameterized by $\phi$ and is used to estimate the state value. The value network is also a 3-layer MLP, which shares the same structure as the policy network. Except for the observation $\mathbf{o}_t$, the privileged observation $\mathbf{s}_t \in \mathbb{R}^{43}$ used by the value network also has access to body position $\bm{p}_t \in \mathbb{R}^3$, depth prior $\bm{d}_t \in \mathbb{R}^{24}$, defined as:
\begin{equation}
    \mathbf{s}_t = \begin{bmatrix} \mathbf{o}_t &  \bm{p}_t & \bm{d}_t  \end{bmatrix}^{T}.
\end{equation}
$\bm{d}_t$ is gained with 3DGS's depth image and is downsampled to the desired size by average pooling.  The value function is not needed at runtime, hence the access to privileged information from the simulator is not a practical limitation.

\subsubsection{Context estimator network}
We incorporated a $\beta$-variational autoencoder ($\beta$-VAE) \cite{higgins2017beta} based CENet~\cite{nahrendra2023dreamwaq} with visual perception from the 3DGS model (training time) or real camera (deployment time). CENet is designed for encoding the drone's surrounding environment, especially the spatial relationship with obstacles, to a latent vector $\mathbf{z}_t$ to enable runtime adaptivity to the environment. CENet is designed with an encoder-decoder structure. The encoder processes the input history observation $\mathbf{o}_t^H$, extracting latent environment representations $\mathbf{z}_t$. The decoder reconstructs the subsequent observation $\mathbf{o}_{t+1}$. The $\beta$-VAE objective consists of both a reconstruction loss and a latent regularization term. The full CENet loss can be expressed as:
\begin{equation}
\small
\mathcal{L}_{CE} = \textrm{MSE}(\hat{\mathbf{o}}_{t+1}, \mathbf{o}_{t+1}) + \beta D_{KL}(q_{\mathbf{z}_t}(\mathbf{z}_t | \mathbf{o}_t^H) || p(\mathbf{z}_t)),
\end{equation}

where $\hat{\mathbf{o}}_{t+1}$ is the reconstructed observation at the next timestep, and $q_{\mathbf{z}_t}(\mathbf{z}_t | \mathbf{o}_t^H)$ is the posterior distribution of the latent variable conditioned on the input observation. The prior distribution $p(\mathbf{z}_t)$ is assumed to follow a standard normal distribution. The backbone of our CENet is a three-layer MLP, sharing the same structure as the policy network. We used a history observation of the last 5 time steps as CENet's input, i.e. $\mathbf{o}_t^H = \mathbf{o}_t^5$.


\subsection{Training Procedure}
\subsubsection{Reward function}
We designed our reward function for two main objectives, (i) smooth and stable dynamic control; (ii) efficient and safe navigation. To enable our method to better transfer to different environments and agents, we tried to avoid finetuning a complicated reward function. The details of our reward function are listed in Table \ref{tab:reward_function}. The total stepwise reward of the policy for taking action at each state is calculated by:
\begin{equation}
    r_t(\mathbf{s}_t, \mathbf{a}_t) = \sum r_i w_i.
\end{equation}

\begin{table}[h]
    \centering
    \small
    \renewcommand{\arraystretch}{1.3}
    \setlength{\tabcolsep}{3pt}
    \caption{Reward function terms and their respective weights. Here, early terminations includes (i) drone's height exceeds the ceiling's height (3m), (ii) drone's linear velocity exceeds the threshold (20m/s), (iii) drone has been out of bound for more than 3m, i.e. $\bm{x}_{o.b.}, \; \bm{y}_{o.b.} \geq 3$, $\bm{q}_0$ is the drone's initial quaternion, $\bm{h}_{\text{target}}$ is the target height of the drone (also serving as the initial hover height), $\hat{\bm{y}}_{\text{yaw}}$ is the normalized heading direction of the drone, $\bm{d}_{\text{wp}}$ is the distance to the next waypoint, $\bm{d}_{\text{obst}}$ is the distance to the closest obstacle in the drone's field of view (FOV), $\bm{x}_{\text{o.b.}}$ and $\bm{y}_{\text{o.b.}}$ are the distances from the 3DGS map boundary, and $\bm{v}_{\text{des}}$ is the desired velocity direction from the pre-planned reference trajectory.}
    \label{tab:reward_function}
    \begin{tabular}{l c c}
        \toprule
        \textbf{Reward} & \textbf{Equation ($r_i$)} & \textbf{Weight ($w_i$)} \\
        \hline
        \multicolumn{3}{c}{\textbf{Safe Control Rewards}} \\
        \hdashline
        Survival & $\notin \{\text{early terminations}\}$ & $8.0$ \\
        Linear velocity & $\|\bm{v}\|^2$ & $-0.5$ \\
        Pose & $\|\bm{q} - \bm{q}_0\|$ & $-0.5$ \\
        Height & $(\bm{h} - \bm{h}_{\text{target}})^2$ & $-2.0$ \\
        Action & $\|\bm{a}_t\|^2$ & $-1.0$ \\
        Action rate & $\|\bm{a}_t - \bm{a}_{t-1}\|^2$ & $-1.0$ \\
        Smoothness & $\|\bm{a}_t - 2\bm{a}_{t-1} + \bm{a}_{t-2}\|^2$ & $-1.0$ \\
        \hline
        \multicolumn{3}{c}{\textbf{Efficient Navigation Rewards}} \\
        \hdashline
        Yaw alignment & $\frac{\bm{v}_{xy}}{\|\bm{v}_{xy}\|} \cdot \hat{\bm{y}}_{\text{yaw}}$ & $0.25$ \\
        Waypoint & $\exp(-\bm{d}_{\text{wp}})$ & $2.0$ \\
        Obstacle avoidance & $\bm{d}_{\text{obst}}$ & $1.0$ \\
        Out-of-map & $\bm{x}_{\text{o.b.}}^2 + \bm{y}_{\text{o.b.}}^2$ & $-2.0$ \\
        Ref. traj. tracking & $\|\frac{\bm{v}}{\|\bm{v}\|} - \frac{\bm{v}_{\text{des}}}{\|\bm{v}_{\text{ref}}\|}\|$ & $-2.0$ \\
        \bottomrule
    \end{tabular}
\end{table}

\subsubsection{Domain randomization}
Domain randomization serves as a standard approach today in robotics reinforcement learning. By changing the environment's appearance, dynamics, and sensor noise, the agent can learn a more robust policy that generalizes to unseen environments and achieves better sim-to-real transitions.
The randomization parameters are listed in Table \ref{table:domain_randomization}.

\begin{table}[h!]
    \centering
    \small
    \caption{Parameter randomization ranges and units.}
    \begin{tabular}{l c c}
        \toprule
        \textbf{Parameter} & \textbf{Randomization range} & \textbf{Unit} \\
        \midrule
        Mass & $[1.0, 1.5]$ & kg \\
        Max. thrust & $[22, 30]$ & N \\
        $I_x, I_y$ inertial & $[0.075, 0.125]$ & $\text{kg}\cdot m^2$ \\
        $I_z$ inertial & $[0.15, 0.25]$ & $\text{kg}\cdot m^2$ \\
        Motor delay factor & $[0.5, 0.8]$ & - \\
        Body rate delay factor & $[0.5, 0.8]$ & - \\
        Drag force coefficient & $[0.4, 0.6]$ & - \\
        \bottomrule
    \end{tabular}
    \label{table:domain_randomization}
\end{table}

\subsubsection{Guiding drone to fly desired trajectories}
\label{sec:nav_reward}

In our reinforcement learning framework for drone navigation, we incorporate multiple reward terms to encourage the drone to fly through obstacles while ensuring collision avoidance. A key component of our approach is defining waypoints and precomputing a reference trajectory based on a point cloud generated from a 3DGS representation of the environment. 

One critical reward component is the \textbf{waypoint reward} \( r_{\text{wp}} \), which encourages the drone to approach the next waypoint on the reference trajectory. The reward is formulated as:
\begin{equation}
r_{\text{wp}} =  \left( e^{-{\|\bm{p} - \bm{w}_{\text{next}}\|^2 }} \right),
\end{equation}
where \( \bm{p} \) is the current drone position, \( \bm{w}_{\text{next}} \) represents the next waypoint (i.e., the closest waypoint with \( x_{\text{wp}} > x_{\text{drone}} \)).

To further reinforce trajectory tracking, we introduce a \textbf{reference trajectory tracking reward} \( r_{\text{traj}} \), which penalizes deviations between the normalized velocity of the drone and the desired velocity along the trajectory:
\begin{equation}
r_{\text{traj}} = \left\| \frac{\bm{v}}{\|\bm{v}\|} - \frac{\bm{v}_{\text{des}}}{\|\bm{v}_{\text{des}}\|} \right\|,
\end{equation}
where $\frac{\bm{v}}{\|\bm{v}\|}$ is the normalized velocity of the drone, $\frac{\bm{v}_{\text{des}}}{\|\bm{v}_{\text{des}}\|}$ is the desired velocity direction drawn from the reference trajectory.

Additionally, to ensure safe navigation, we incorporate an \textbf{obstacle avoidance reward} \( r_{\text{obss}} \), which encourages the drone to maintain a safe distance from obstacles visible in its FOV. The reward formulation is as follows:
\begin{equation}
r_{\text{obst}} = \bm{d}_{\text{obst}}, \quad \bm{d}_{\text{obst}} < \bm{d}_{\text{th}},
\end{equation}
where \( \bm{d}_{\text{obss}} \) represents the minimum distance to the nearest obstacle in drone's FOV, \( \bm{d}_{\text{th}} \) is the predefined threshold for proximity, which is set to 0.5m in our method. 

These reward components contribute to the overall reward function, which balances waypoint tracking, trajectory adherence, and obstacle avoidance to optimize drone navigation in complex environments. The detailed reward factors $\omega_i$ for each term are listed in \ref{tab:reward_function}.

\subsubsection{Curriculum training for generalizable navigation policy}
\label{sec:curriculum_training}
Beyond training a single policy for a long horizon trajectory, our method can also train generalizable policies that can adapt to different surrounding environments and conduct safe navigation in them. We achieve end-to-end generalizable gate position detection and safe navigation with a curriculum training method. We have three different 3DGS environments for training, each one with a unique gate position. By rolling training across these different environments, we finally trained a policy that can adapt to different gate positions and achieve generalizable navigation. The reward function (Table. \ref{tab:reward_function}) is kept the same during training. We trained the policy in each environment for 5 times, 100 epochs per time; we return the learning rate to the initial value during every environment transition, other hyperparameters are listed in Table \ref{table:hyper-param}.


\section{Experimental Results}

\subsection{Training efficiency comparison with other methods}
\begin{figure}[h!]
    \centering
    \includegraphics[width=\linewidth]{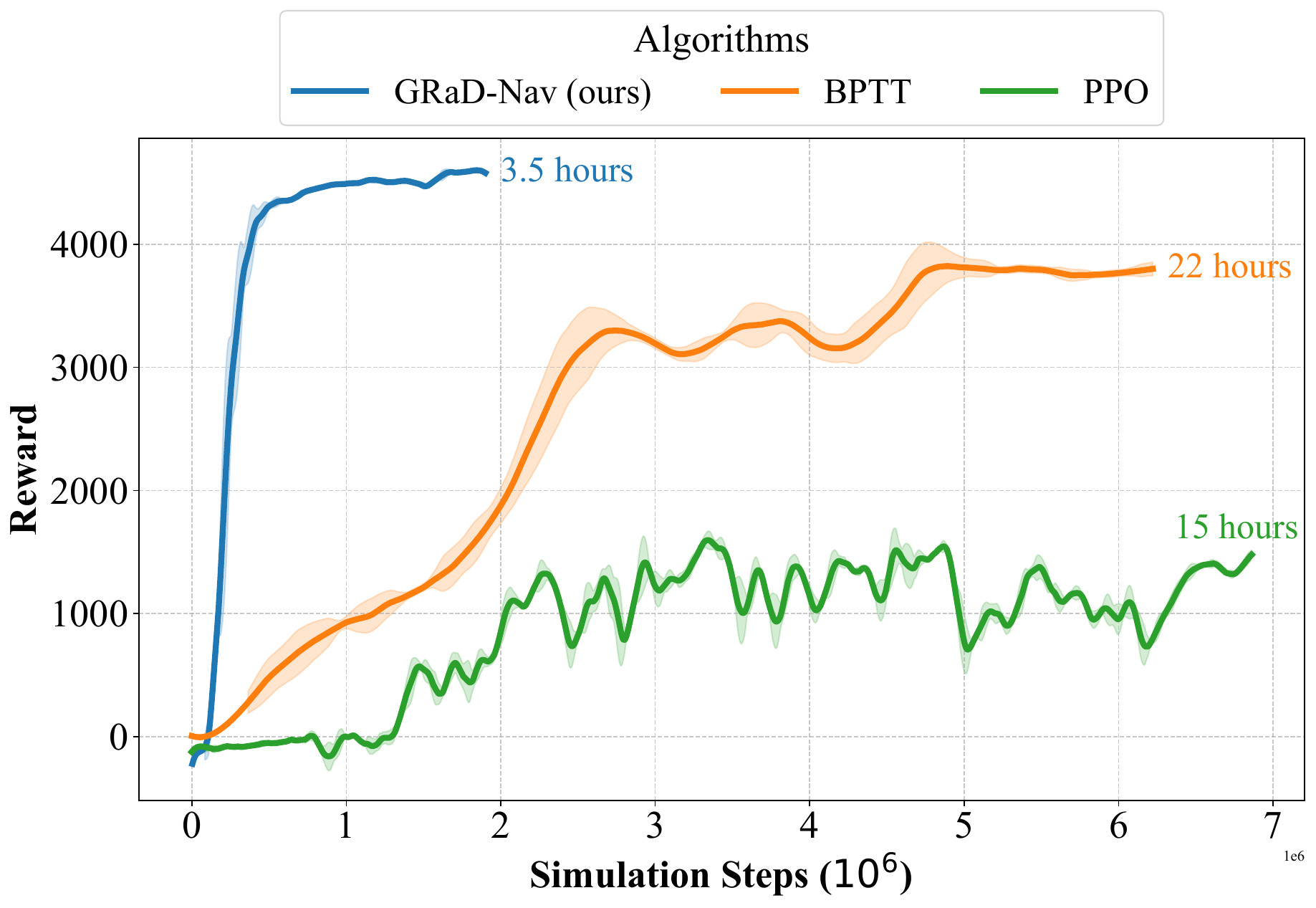}
    \caption{Sample efficiency and wall-clock time comparison benchmark of different algorithms on drone's vision-based end-to-end navigation policy training.}
    \label{fig:algo_benchmark}
\end{figure}

\begin{figure*}[t]
    \centering
    \includegraphics[width=0.9\linewidth]{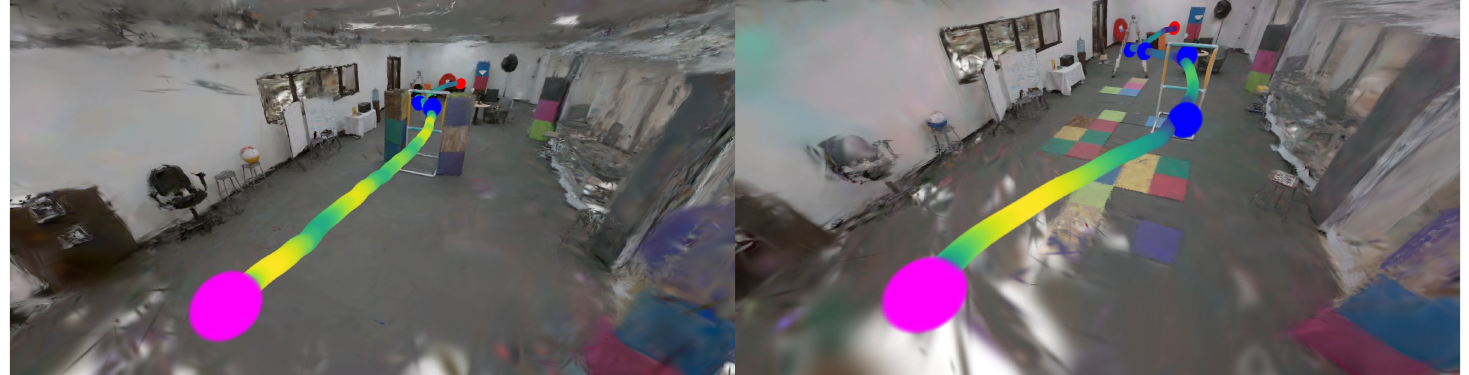}
    \caption{Example success trajectories in hybrid simulation environments achieved by the proposed method. The left one is ``middle gate" and the right one is ``right gate", aligning with the ablation study Table \ref{table:ablation_test}.}
    \label{fig:sim_traj}
\end{figure*}

To validate GRaD-Nav's training efficiency, we compared our method with other differentiable and non-differentiable RL algorithms. Back Propagation Through Time (BPTT)~\cite{mozer2013focused} is a critic-free differentiable reinforcement learning algorithm that passes gradients through the whole trajectory; Proximal Policy Optimization (PPO)~\cite{schulman2017proximal} is a common non-differentiable, online reinforcement learning algorithm. All three algorithms were used for training policies that navigate through a trajectory of around 20m, passing through several challenging unstructured obstacles, which can be seen in \ref{fig:sim_traj}. As much as possible, we used identical hyperparameters (Table \ref{table:hyper-param}) for different algorithms. It is to be noted that BPTT samples the whole trajectory for policy updating, meaning the horizon length equals the episode length, which can take up a lot of memory for each environment and heavily limits its maximum allowed number of parallel simulated environments. The experiment results in Fig.\ref{fig:algo_benchmark} show that (i) non-differentiable RL can struggle to train a satisfactory policy for this end-to-end visual navigation task within $1 \times 10^7$ samples; (ii) compared to vanilla BPTT, our method achieves over 300\% sample efficiency and uses only 20\% of training time to deliver a better policy.

\subsection{Ablation study of our methods}
\label{sec:ablation_study}
To validate that each module of our method is not redundant but necessary for safe navigation, and to determine each module's contribution to the entire training framework, we conduct a comprehensive ablation test to train the policy networks with certain modules ablated on two different surrounding environments. We train each policy with the same reward function as in Table \ref{tab:reward_function} and the same hyperparameters setting as Table \ref{table:hyper-param} for 600 epochs. Our ablation test metrics include: (i) training reward, (ii) test reward, and (iii) test success rate. Training reward is the highest reward policy achieved during training time. After training, we roll out the best policy and conduct simulations for 10 drones in parallel. The simulated drone's initial positions are randomized in a cube with side length of 1m, centered at (0,0,1.3)m; their initial poses are also randomized, $\phi, \theta, \psi \in [-0.25,0.25].$ Other dynamic properties of the simulated drones are also randomized as in Table \ref{table:domain_randomization}. The assessments of a success trajectory include: (i) not suffering from early termination as discussed in Table \ref{tab:reward_function}; (ii) getting close enough (${\|\bm{p} - \bm{w}_{\text{next}}\|^2 }_{min} \leq 0.3 \text{m}$) to each waypoint in sequence; (iii) keeping a safe distance ($\geq$ 0.2m) to any obstacles on the trajectory.

\begin{table*}
\centering
    \small
    \caption{Simulation results of ablation test, the reward function is defined in Table \ref{tab:reward_function}, each method is trained with GRaD-Nav algorithm and hyperparameters list in Table \ref{table:hyper-param}. Example trajectories are shown in Fig. \ref{fig:sim_traj}.}
    \begin{tabular}{ccccccc}
    \toprule 
    \textbf{Trajectories} & \multicolumn{3}{c}{\textbf{Long traj. middle gate}}     & \multicolumn{3}{c}{\textbf{Long traj. Right gate}}      \\
    \cmidrule(r){1-1} \cmidrule(lr){2-4} \cmidrule(rr){5-7} 
    Methods               & Training reward & Evaluation reward & Success rate & Training reward & Evaluation reward & Success rate \\
    \midrule 
    w/o visual obs.       & 3720.1          & 3761.4  $\pm$ 60.1    & 0/10         &    4162.4             &    4148.3    $\pm$ 76.3     &   0/10           \\
    w/o RGB; w/ depth     &   3828.6              & 3805.7  $\pm$ 45.9          &   0/10           &       4168.2          &  4161.7 $\pm$ 27.5          &   0/10           \\
    w/o velocity          & 3579.6                & 3430.1 $\pm$ 116.8            & 0/10              &      4030.1           &    4012.0   $\pm$ 113.4      &       0/10       \\
    w/o CENet             & 4014.0          & 4124.3 $\pm$ 140.9     & 4/10         & 4641.7           &    4583.9  $\pm$ 238.1        &     5/10         \\
    Proposed              & \textbf{4622.1}          & \textbf{4893.6}  $\pm$ 24.8    & \textbf{8/10}         & \textbf{4928.5}         & \textbf{4718.3}  $\pm$ 12.6    & \textbf{7/10} \\      
    \bottomrule 
    \end{tabular}
    \label{table:ablation_test}
\end{table*}

The experiment results show that our proposed method achieves the highest training and evaluation rewards as well as success rate on both trajectories among all methods. As visual perception is our navigation policy's major sensor input, it is not surprising that the policy without visual observation cannot conduct successful navigation. Using depth image rather than RGB image also demonstrated a low success rate in vision-ablated experiments, which emphasized the necessity of accessing RGB information for effective navigation on our task. As introduced in Section \ref{sec:nav_reward}, drone's body linear velocity plays an important role in our policy's navigation-related reward items, velocity-ablated policies would suffer from serious compounding error, and thus cannot fly an ideal trajectory. Without CENet, our method can still train a policy network that achieves high rewards compared to other ablation cases. CENet ablated policies can complete a trajectory successfully when the initial condition is close to the "neutral" initial condition (e.g. position = (0,0,1.3)m).
However, policies without CENet can be vulnerable to any environmental perturbations, which seriously harm the robustness of the policy networks. All of the failure cases without CENet on two trajectories ``crash" due to unsuccessful obstacle avoidance. 

\subsection{Sim-to-real transfer of generalizable policy}
\begin{figure}[h!]
    \centering
    \includegraphics[width=\linewidth]{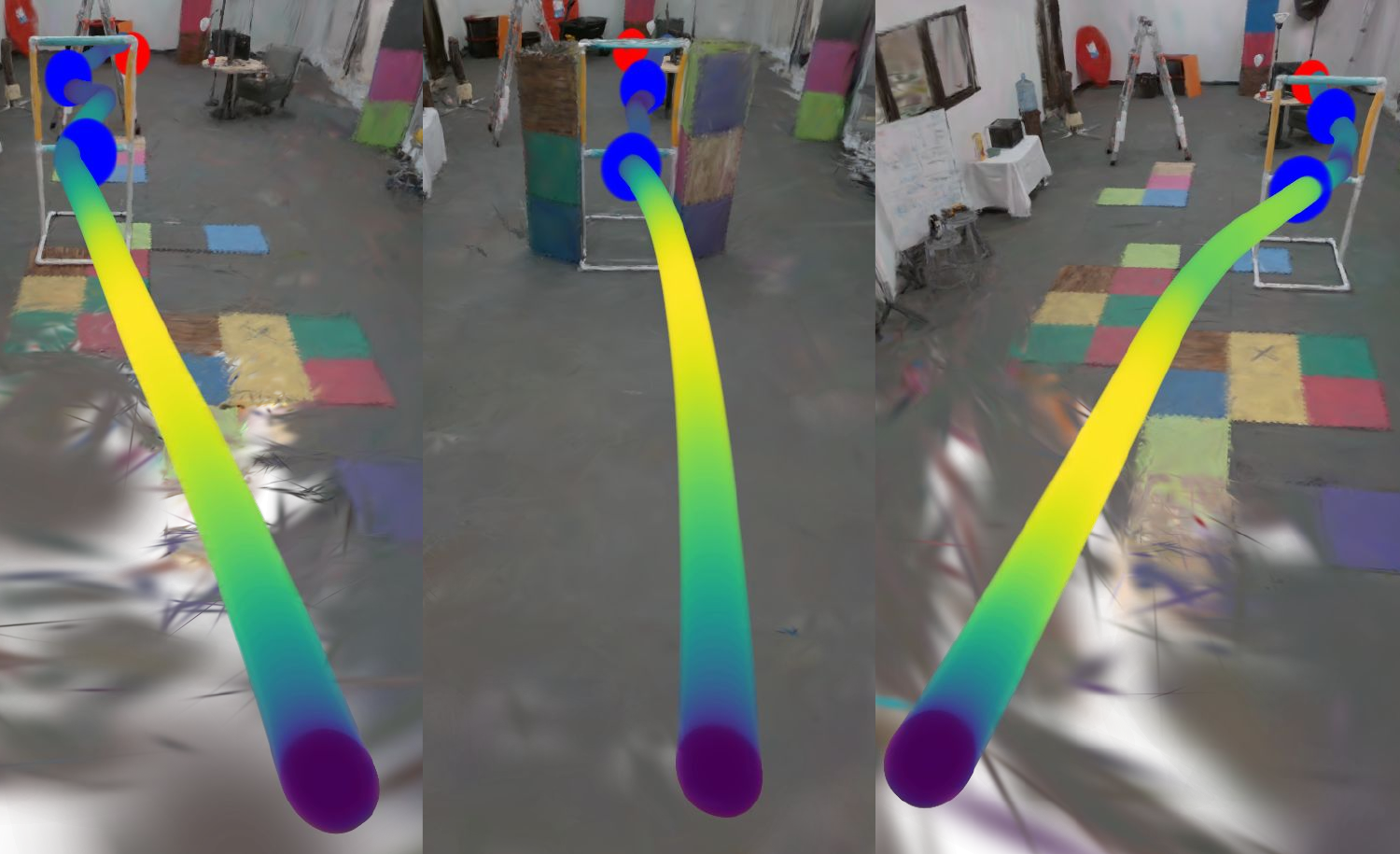}
    \caption{A generalizable policy flying through gates positioned at varying locations with diverse distractor objects in the scene. The colored trajectory represents the drone's velocity, with purple indicating low speed and yellow indicating high speed. }
    \label{fig:gen_traj}
\end{figure}

After training our policy as discussed in Section~\ref{sec:curriculum_training}, we rolled out the final policy and tested it in three different environments separately. Without any prior environmental-related information like the map's name, gate position, or reference trajectory, the policy can only access the first person perspective RGB information rendered from 3DGS. During experiments, the rolled-out policy could "recall" what it learned in different surrounding environments and guide the drone to fly through the gate that is placed at different locations. Fig.~\ref{fig:gen_traj} demonstrates the generalizable policy's variant trajectories in different environments.

\begin{figure}[h!]
    \centering
    \includegraphics[width=\linewidth]{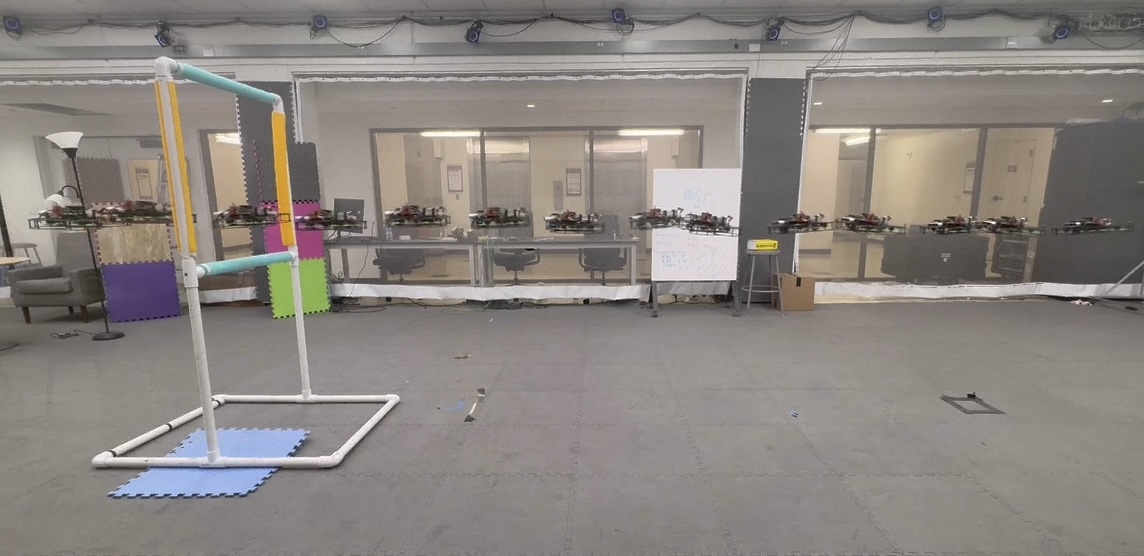}
    \caption{Robot hardware experiments of drone flying through middle gate.}
    \label{fig:real_traj}
\end{figure}

 Thanks to the high fidelity first person perspective RGB image rendered with 3DGS as showed in Fig.~\ref{fig:first_view_compare}, we can conduct convenient zero-shot sim-to-real transfer. Our drone is mounted with a Pixracer low-level flight controller, a Nvidia Jetson Orin Nano onboard computer, and an Intel Realsense D435 camera. The onboard policy inference frequency is 30 Hz. Fig.~\ref{fig:real_traj} is a demonstration of the successful trajectories navigating through gates with variant y-positions. Fig.~\ref{fig:first_view_compare} further validates CENet’s role in environment understanding. We recorded the latent vector $\mathbf{z}_t$ at three flight stages—approaching, passing through, and after the gate—and applied PCA. The latent vectors during passage show a more compact distribution than the other stages. By comparing three methods' real robot test performance in Table~\ref{tab:real-robot_success_rate}, we can conclude that (i) the sim-to-real gap of our method is reasonably low; (ii) CENet is essential in delivering robust sim-to-real transfer; (iii) only using depth data in real robot deployment would lead to a severe loss of environment detail thus seriously harming the success rate.

 \begin{figure}[h!]
    \centering
    \includegraphics[width=\linewidth]{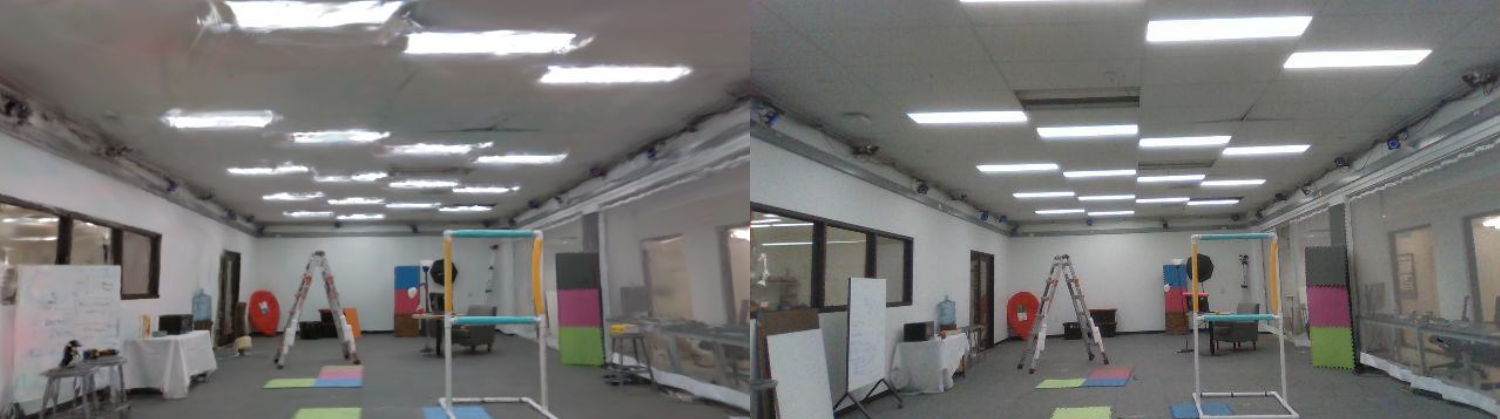}
    \caption{Comparison on drone's first person perspective image rendered with 3DGS in simulator (left) and captured with Intel Realsense D435 camera in real robot deployment (right).}
    \label{fig:first_view_compare}
\end{figure}

\begin{figure}[h!]
    \centering
    \includegraphics[width=\linewidth]{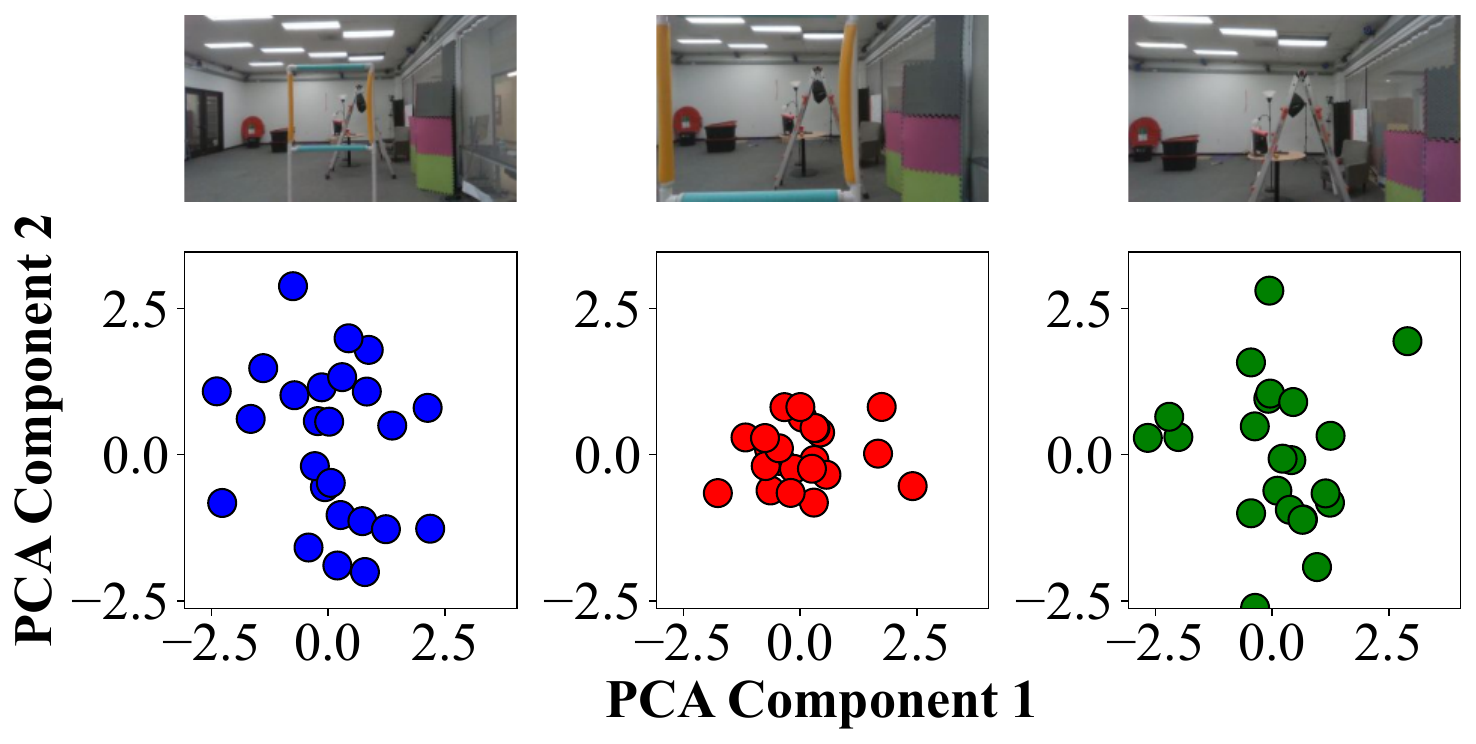}
    \caption{The drone's first-person views at three different stages during the real-robot experiment, along with the PCA visualization of the latent vector $\mathbf{z}_t$ from the context encoder CENet at each stage.}
    \label{fig:first_view_compare}
\end{figure}

\begin{table}[t]
\centering
\scriptsize
\caption{Experimental results of generalizable policies trained using different methods. In simulation experiments, the drone's randomized initialization settings and the evaluation metrics for successful trajectories are consistent with those in Section~\ref{sec:ablation_study}. In real-world experiments, the drone was initialized under the same conditions for each test, and success was determined by whether it flew through the gate without collision.}
\begin{tabular}{cccc}
\toprule
\multirow{2}{*}{\textbf{Methods}} & \multicolumn{3}{c}{\textbf{Success rate (sim. \textbar\ real)}}     \\
                                  & Left gate     & Middle gate   & Right gate    \\ \midrule
w/o RGB; w/ depth                 & 4/10 \textbar\ 1/10          & 7/10 \textbar\ 1/10          & 5/10 \textbar\ 0/10               \\
w/o CENet                         & 9/10 \textbar\ 0/10          & \textbf{10/10} \textbar\ 2/10          & 7/10 \textbar\ 1/10          \\
Proposed                          & \textbf{10/10 \textbar\ 7/10} & \textbf{10/10 \textbar\ 7/10} & \textbf{9/10 \textbar\ 6/10} \\ \bottomrule
\end{tabular}
\label{tab:real-robot_success_rate}
\end{table}

\section{Conclusions}
In this paper, we present a novel framework that integrates 3DGS with DDRL to train a vision-based drone navigation policy. By leveraging high-fidelity 3D scene representations and differentiable simulation, our approach enhances sample efficiency and sim-to-real transfer. Experimental results demonstrate that our method outperforms existing approaches in training efficiency. Our method also demonstrates robustness and generalization ability. 

\emph{Limitations:} Our method relies on hand-crafted reward shaping (e.g., trajectory waypoints), limiting it to single-task execution like gate traversal. It also depends on reliable velocity estimates from VIO, which may be inconsistent. Future work includes (i) multi-task training with language input, (ii) improving generalization via stronger backbones and diverse environments, and (iii) extending to contact-rich tasks such as mobile or aerial manipulation.
\section{Acknowledgment}
The authors would like to thank Chenghao Zhu, Keiko Nagami and Javier Yu for fruitful discussion on FiGS. The GRaD-Nav algorithm was developed based on SHAC's open-sourced codebase \cite{xu2022accelerated}, $\beta$-VAE based CENet code was referred to \cite{DreamWaQ_repo}.

\newpage 
\bibliographystyle{IEEEtran}
\bibliography{ref}
\section{Appendix} \label{sec:appendix}
\setlength{\tabcolsep}{3pt}

\begin{table}[h]
    \centering
    \scriptsize
    \caption{Hyper-parameters table of different training methods demonstrated in Fig.~\ref{fig:algo_benchmark}.}
    \label{table:hyper-param}
    \begin{tabular}{lccc}
        \toprule
        \textbf{Parameters} & \textbf{GRaD-Nav} & \textbf{PPO} & \textbf{BPTT} \\
        \midrule
        Number of envs             & 128   & 128   & 32  \\
        Episode length             & 600   & 600   & 600  \\
        Discount factor $\gamma$   & 0.99  & 0.99  & 0.99  \\
        Actor learning rate        & 1e-4  & 1e-4  & 1e-4  \\
        Critic learning rate       & 1e-4  & 1e-4  & -  \\
        CENet learning rate        & 5e-4  & 5e-4  & 5e-4  \\
        GAE $\lambda$              & 0.95  & 0.95  & 0.95  \\
        Horizon length             & 32    & 32    & 600  \\
        Critic updates             & 16    & -     & 16  \\
        Clipping parameter $\epsilon$  & - & 0.1   & -  \\
        Entropy coefficient        & -     & 1e-3  & -  \\
        \bottomrule
    \end{tabular}
\end{table}

\end{document}